\documentclass[letterpaper, 10 pt, conference]{ieeeconf}  

\IEEEoverridecommandlockouts                              

\usepackage{graphics} 
\usepackage{epsfig} 
\usepackage{mathptmx} 
\usepackage{times} 
\usepackage{amsmath} 
\usepackage{amssymb}  
\usepackage{multirow}
\usepackage{breqn}

\title{\LARGE \bf
How Does Perception Affect Safety: New Metrics and Strategy
}

\author{Xiaotong Zhang, Jinger Chong and Kamal Youcef-Toumi
\thanks{All authors are with the Mechatronics Research Laboratory, Massachusetts Institute of Technology, Cambridge, MA, 02139, USA
{\tt\small \{kevxt, jinger, youcef\}@mit.edu }}
\thanks{This research was made possible by the support and partnership
of King Abudlaziz City for Science and Technology
(KACST) through the Center for Complex Engineering
Systems at Massachusetts Institute of Technology (MIT) and
KACST.}
}

\begin{document}

\maketitle
\thispagestyle{empty}
\pagestyle{empty}

\begin{abstract}

Perception serves as a critical component in the functionality of autonomous agents. However, the intricate relationship between perception metrics and robotic metrics remains unclear, leading to ambiguity in the development and fine-tuning of perception algorithms. In this paper, we introduce a methodology for quantifying this relationship, taking into account factors such as detection rate, detection quality, and latency. Furthermore, we introduce two novel metrics for Human-Robot Collaboration safety predicated upon perception metrics: Critical Collision Probability (CCP) and Average Collision Probability (ACP). To validate the utility of these metrics in facilitating algorithm development and tuning, we develop an attentive processing strategy that focuses exclusively on key input features. This approach significantly reduces computational time while preserving a similar level of accuracy. Experimental results indicate that the implementation of this strategy in an object detector leads to a maximum reduction of 30.091\% in inference time and 26.534\% in total time per frame. Additionally, the strategy lowers the CCP and ACP in a baseline model by 11.252\% and 13.501\%, respectively. The source code will be made publicly available in the final proof version of the manuscript.
\end{abstract}


\section{Introduction}
Intelligent agents and machines, such as robots, autonomous vehicles, and machines, are becoming increasingly important in human daily life and industrial applications \cite{zhang2019design,zhang2022magnetohydrodynamic,zhang2022systematic}. 
They require perception algorithms to sense and understand the dynamic environment fast and accurately, dynamic planning algorithms to proactively update the plan, and control algorithms to generate the control commands to the actuators. This is the so-called pipeline frameworks for autonomous agents.
Perception is an important but computationally expensive component. 

However, an intensive literature review shows that several issues and shortcomings still exist. First, the relationship between the metrics of perception tasks and the metrics of the robot tasks is unclear. Researchers working on a specific perception problem focus on improving the metrics for the specific perception task. However, the ultimate goal of perception algorithms is for real-world applications. Thus, a model, that is beneficial for understanding the effect of the perception performance on robot performance, is desired and necessary. Second, the accuracy-speed-tradeoff is an inevitable issue when developing and tuning perception models. Neither a perfectly accurate with a slow inference speed nor a fast algorithm with a very low accuracy is desired for autonomous agents. However, there is no model for understanding this problem and guidelines for finding the optimal balance points when tuning models. Third, many papers and benchmarks focused and reported on the accuracy aspect of metrics, such as classification accuracy, precision, recall, F-1 score, AUC-ROC, etc., with no mention to the speed metrics. We would like to argue, in this paper, that the speed aspect is crucial in applications which require real-time decision making and reaction to the dynamic environment, such as robotics, autonomous vehicles, remote surgery, gaming, monitoring and surveillance, etc. 




In order to mitigate or address the above-mentioned issues, we develop and propose a methodology to model the robot performance metrics based on perception task metrics. We first separate and decouple the perception metrics into three categories, namely the algorithm's detection rate, detection quality, and latency. Then, we investigate and discuss their influence on robot safety metrics separately. Here, we take the safety issues in Human-Robot Collaboration (HRC) tasks as an example. We further propose two new metrics for the HRC tasks, called Critical Collision Probability (CCP) and Average Collision Probability (ACP), which link the perception metrics to the safety metrics of the robot in real-world deployment. We preliminarily validate our modeling with the analysis of the attentive processing strategy. The trends of the model agree well with realistic situations. 

To demonstrate the usefulness of the modeling and the metrics, we further develop and propose a new strategy to improve the safety of HRC benefiting from our models by enhancing the inference speed of the algorithm while maintaining similar accuracy, which is inspired by how human brains process information. One essential difference between human beings and machines is that human beings weigh the priority and importance of different tasks and inputs, process the essential components with more resources, and dynamically adjust the focus based on task characteristics or other guidance. This mechanism has been studied extensively and verified with examples, e.g. foveal-peripheral vision \cite{xing2000center,vater2022peripheral}, inattentional blindness \cite{mack2003inattentional}, change blindness \cite{simons1997change}, etc.

Inspired by these functions and their effectiveness in human perception, it would be desirable for machines to selectively process the essential components determined by the characteristics and states of the tasks and surrounding agents. Another inspiration of this attentive processing method is the nature of visual inputs that only a small portion of the visual inputs are related to the task or the short-term decision making. 
Thus, respecting this nature can dramatically decrease the computation resources required, accelerating the model inference or reducing the hardware requirements and energy consumption for large model inference. 


To this end, we develop and propose a new generic strategy applicable to most computer vision tasks, i.e. attentive processing. In this strategy, the algorithm focuses on and processes only the essential components related to the task with an ensemble of neural networks with different scales. Firstly, we dynamically update the attentive region based on the results from previous frames, respecting the continuity of agent states in real-world applications. Secondly, only the local information within the attentive region is fed into the neural network with the corresponding scale to reduce unnecessary processing in a larger-scale network. Thirdly, the processing of the attentive region results is mapped back to the global input frame. Then, information from the current frame is processed with a much more efficient and lightweight neural network instead of processing every pixel in the global frame with a denser and deeper neural network. The results of the current frame can guide the selection of the attentive region for the next frame. This strategy can be used along with most computer vision algorithms as an additional layer to select only the essential components for processing. In this paper, we use a video object detection task as a case study for verification and validation purposes.

To summarize, the contributions of this work are four-fold: (i) A methodology for modeling robot performance metrics based on perception metrics. Here, we use HRC safety metrics considering object detection recall, IoU, and latency, as a case study. (ii) We model and propose two novel comprehensive perception metrics considering speed-accuracy-tradeoffs, namely ACP and CCP, in the context of HRC tasks. (iii) A new generic attentive processing strategy on top of efficient learning methods to select, track, and update the essential components in input information for reduced computation resource requirement and effort by utilizing the sparsity in input data. Experiments verify that the attentive processing strategy can enhance safety and speed in human-robot collaboration. (iv) We demonstrate the importance of focusing on real-time algorithms and considering speed-accuracy-tradeoffs in real-world applications.







\section{Related Work}
To the best of our knowledge, our work is very unique. Some works related to ours are summarized in the following sub-sections.

\subsection{Perception and Robot Metrics}
Investigation and analysis of the relationship between perception and robot metrics are crucial. Some previous works analyze the relationship between latency and robot performance \cite{handa2012real,844125,behnke2004predicting,sermanet2007speed}. The most related work to ours is \cite{8636976}, which modeled and analyzed the influence of visual latency on high-speed sense and avoidance in a drone application. This work also demonstrated and verified the effectiveness of event cameras in high-speed applications. However, this work doesn't include detection rate and detection quality in the analysis, and it poses strong assumptions (e.g. unchanged longitudinal velocity) in the analysis, which are difficult to generalize to other applications. Our work is more comprehensive, including detection rate and quality, and suitable for generic collision avoidance problems.

\subsection{Efficient Deep learning}
Accelerating and decreasing the computation resource requirements of model inference is an active research area. Methods, including quantization \cite{vanhoucke2011improving,dong2019hawq,guo2018survey,micikevicius2017mixed}, deep compression \cite{han2017efficient}, knowledge distillation \cite{gou2021knowledge}, sparsification \cite{zhou2021effective}, and pruning \cite{lecun1989optimal,molchanov2016pruning,blalock2020state,han2015learning}, have shown promising results in the application of edge devices with limited computational resources. These methods tweak the architecture and the precision of the neural networks themselves to remove unnecessary connections and components, reducing the precision of weights and activations, and/or replacing the large network with a more efficient network. However, none of these utilizes the sparsity of the input features. 
Our methods can work along with these efficient deep learning methods by distilling only important features for processing.

\section{Safety Metric Modeling}



\subsection{Problem Setup}
\begin{figure}[!t]
\begin{center}
\includegraphics[width=0.7\columnwidth]{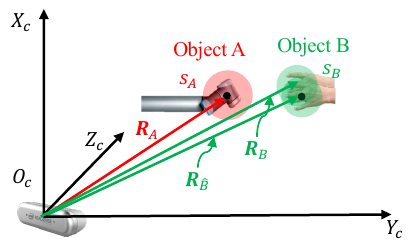}
\end{center}
\caption{Problem setup with camera coordinates and necessary variables for analysis.}
\label{fig: coordinates}
\end{figure}

The problem setup is shown in Figure \ref{fig: coordinates}. In this model, we consider a general HRC problem, where the human subjects focus on their own tasks, and the robot facilitates by conducting some other auxiliary tasks. The robot executes the planned path until a human is detected within a safety range by an object detector or human pose estimator. After the detection, the robot either re-plans the path or stops in urgent situations. In this model, for simplicity, we assume that the robot slows down after detecting, and speeds are assumed constant during the frames considered. Our future work will consider the more complex situations with arbitrary motion of human hands and robots. 

Without loss of generality, we assume that the camera is fixed in the world coordinates. Thus, all variables are denoted in the camera coordinates. We denote the gripper as object A with a spherical safety margin $s_A$ and the hand as object B with a spherical safety margin $s_B$. The parameter $s_A$, or $s_B$, is the radius of the safety margin. Here, we assume that the safety margins are the same as their sizes. For the human hand, which is detected by perception algorithms, we distinguish the ground truth $B$ and its estimate $\hat{B}$ to clearly analyze the effect of detection quality. The location of robot A can be derived through encoders and kinematics, whose error is smaller. Here, we only distinguish $B$ and $\hat{B}$ for analysis of detection quality, and use $\hat{B}$ to represent the location of $B$.

\begin{figure}[!t]
\begin{center}
\includegraphics[width=0.68\columnwidth]{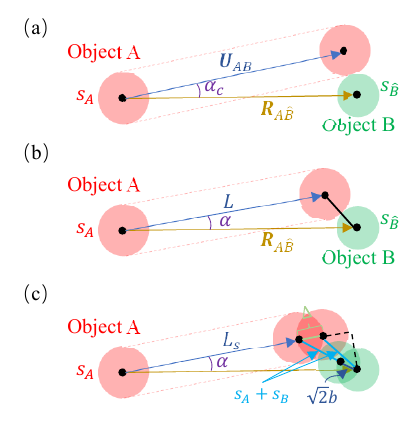}
\end{center}
\caption{The illustration of collision condition. (a) critical condition; (b) general condition. (c) after considering the detection shift.}
\label{fig: collision}
\end{figure}

In the camera coordinates, the position vectors of A and $\hat{B}$ are $\mathbf{R}_A$ and $\mathbf{R}_{\hat{B}}$, respectively, with velocity vectors $\mathbf{U}_A = \dot{\mathbf{R}}_A$ and $\mathbf{U}_{\hat{B}} = \dot{\mathbf{R}}_{\hat{B}}$. The position vector from A to B is $\mathbf{R}_{A{\hat{B}}} = \mathbf{R}_{\hat{B}}-\mathbf{R}_A$. The distance between A and B can be calculated as $d_{A{\hat{B}}} = |\mathbf{R}_{A{\hat{B}}}|$. Collision in this paper is defined when the condition $d_{A{\hat{B}}} < s_A+s_B$ is true. The velocity of A relative to B can be calculated as $\mathbf{U}_{A{\hat{B}}} = \mathbf{U}_A - \mathbf{U}_{\hat{B}}$. The illustration of variables to calculate collision condition after being projected to the plane containing $\mathbf{U}_{A{\hat{B}}}$ and $\mathbf{R}_{A{\hat{B}}}$ is shown in Figure \ref{fig: collision}. The angle between $\mathbf{R}_{A\hat{B}}$ and $\mathbf{U}_{A{\hat{B}}}$ is defined as $\alpha$. 
The critical condition can be re-written as $\alpha_{c} = \arctan\left((s_A+s_B)/\sqrt{|\mathbf{R}_{A\hat{B}}|^2-(s_A+s_B)^2}\right)$.
The collision condition can only happen when $\alpha \in[-\alpha_{c},\alpha_{c}]$. Under this condition, the maximum safe travel distance $L$ without considering the bounding box shift can be calculated as follows

\begin{equation}
L = |\mathbf{R}_{A{\hat{B}}}|\cos{\alpha}-\sqrt{(s_A+s_B)^2-|\mathbf{R}_{A{\hat{B}}}|^2\sin^2{\alpha}}
\label{eq: Ls}
\end{equation}

\subsection{Effect of Latency}
The total latency per frame  $T_{t}$ from sensory input to successful response can be decomposed as follows 

\begin{equation}
T_{t} = T_{p} + T_{r}
\end{equation}

where $T_{p}$ and $T_{r}$ are the latency for perception and response, respectively. $T_r$ affects the allowable traveling distance before collision. The effect of $T_p$ is reflected in the number of possible frames to detect humans before collision.


\subsection{Effect of Detection Rate}
The detection rate considers that the algorithms generate false negatives, which can be fatal. In perception tasks, for one frame, the probability of successful detection equals recall of the algorithm, which means $P_{d} = Recall$. We assume that frames are independent. Then, we define a random variable $k$ as the average number of frames needed for the algorithm to make a successful detection. $k$ follows a Geometric Distribution as follows:

\begin{equation}
k \sim Geo(P_{d} = recall)
\end{equation}





\subsection{Effect of Detection Quality}
Detection quality metrics in perception include Intersection over Union (IoU), Mean Average Precision (mAP), F1 score, etc. Here, we use IoU as the detection quality metric. 
We assume that the camera is homogeneous in its imaging plane $o'x'y'$ without skewness and distortion. Because of the linear operation in the camera imaging model and the dimensionless property of IoU, IoU for the object on the image plane equals IoU of the object in the world coordinates. Thus, according to the IoU in the image plane, we can derive the distance between the real object and the predicted bounding box in the world coordinates.




 The $IoU$ under these assumptions is calculated as follows

\begin{equation}
    IoU = \frac{(s_B-b)^2}{2s_B^2-(s_B-b)^2}
    \label{eq:iou}
\end{equation}

where $b$ is the shift between the real location of the object and the predicted location in the world coordinates in the direction of $x$ or $y$. We can derive the value of $b$ from (\ref{eq:iou}) as

\begin{equation}
    b=s_B\left(1-\frac{\sqrt{2IoU}}{\sqrt{IoU+1}}\right)
    \label{eq:b}
\end{equation}

Thus, the total maximum shift in the world coordinate is $\sqrt{2}b$. According to the geometrical relationship shown in Figure \ref{fig: collision}(c), the safe traveling distance decreases by $\Delta$, which is the shift of $A$ along $L$ because of the detection shift of $B$. Here, we consider the maximum effect of $b$ on $L$. Thus, we derive the maximum safe traveling distance $L_s$ as

\begin{dmath}
L_s = max\left(0,L - \left(\sqrt{(s_A+s_B+\sqrt{2}b)^2-|\mathbf{R}_{A\hat{B}}|^2\sin^2{\alpha}}
-\sqrt{(s_A+s_B)^2-|\mathbf{R}_{A\hat{B}}|^2\sin^2{\alpha}} \right)-|\mathbf{U}_{A\hat{B}}|T_r \right)
\label{eq:Ls}
\end{dmath}


\subsection{Safety Metrics}
After considering all the factors discussed above, we could derive the probability of collision $P_c$ as follows:

\begin{equation}
P_{c} = \begin{cases}
P(k>m) = (1-P_d)^m, &\text{if } \alpha\in[-\alpha_c,\alpha_{c}]\\
0, &\text{otherwise}
\end{cases}
\end{equation}

 where $m$ is the number of frames processed by the perception algorithm before traveling the distance $L_s$ and calculated as follows:
 
\begin{equation}
m = \left\lfloor \frac{L_s}{|\mathbf{U}_{AB}|\cdot T_{p}}\right\rfloor
\label{eq: m}
\end{equation}

A larger $m$ means more input frames are available to the algorithm before collision.

\subsubsection{Critical Collision Probability $\mathbf{CCP}$}
Critical Collision Probability $\mathbf{CCP}$ is an important metric, which represents the average collision probability under critical and dangerous conditions. CCP is defined as 

\begin{equation}
\mathbf{CCP} = \mathbb{E}_{\alpha, |\mathbf{R}_{A\hat{B}}|, |\mathbf{U}_{A\hat{B}}| \in C }[P_c]
\end{equation}

where $\mathbb{E}$ is the expectation operator, $C$ is the region of critical conditions. In this condition, $\alpha$, $|\mathbf{R}_{A\hat{B}}|$, and $|\mathbf{U}_{A\hat{B}}|$ may be correlated. An example of the parameter selection is discussed in the Experiments section.

\subsubsection{Average Collision Probability $\mathbf{ACP}$}

Further, we define another safety metric Average Collision Probability $\mathbf{ACP}$, which is the expectation of collision probability over applicable parameter space representing the average performance over a wide range of conditions, as

\begin{equation}
\mathbf{ACP} = \mathbb{E}_{\alpha, |\mathbf{R}_{A\hat{B}}|, |\mathbf{U}_{A\hat{B}}| \in D }[P_c]
\end{equation}

where $D$ is the parameter space for the application. The value of ACP represents the average collision probability under all robot operation conditions.




The two metrics can be considered separately or jointly with weight either as an evaluation protocol of perception algorithms or as a training objective for perception algorithms. 


\begin{figure*}[!t]
\begin{center}
\includegraphics[width=0.93\linewidth]{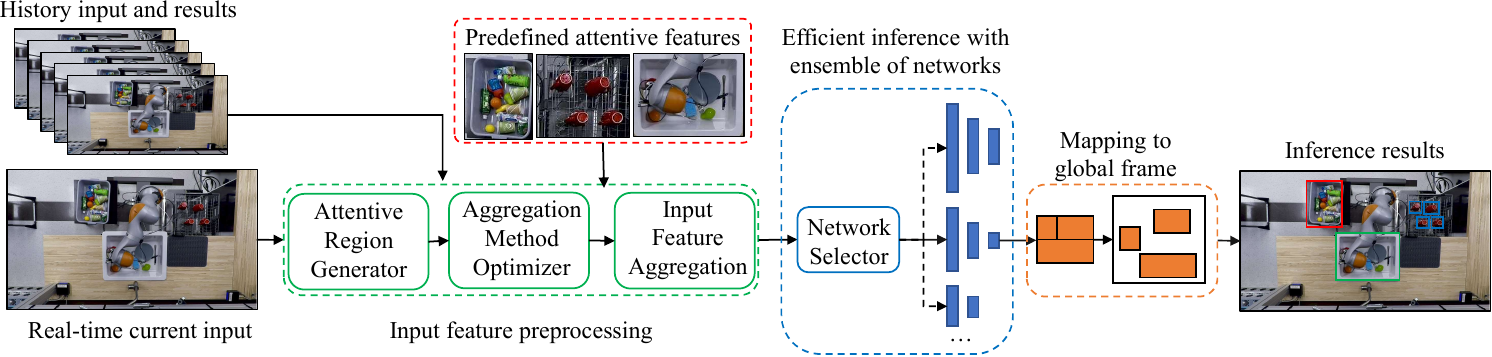}
\end{center}
   \caption{The proposed attentive processing strategy. The essential and attentive features in the real-time current input $\mathbf{I}$ are selected, cropped, and aggregated into a new feature map $\mathbf{I'}$. Because the dimension of $\mathbf{I'}$ is smaller than $\mathbf{I}$, we use the optimal networks within an ensemble of networks to process $\mathbf{I'}$ efficiently with similar performance accuracy. Finally, the prediction results are mapped back to the global input frame.}
\label{fig:method}
\end{figure*}

\section{Attentive Processing Strategy}
The working principle of the proposed attentive processing approach is shown in Figure \ref{fig:method}. The main components of the framework can be decomposed into Input Feature Preprocessing (IFP), Efficient Inference (EI), and Result Mapping (RM). We use video object detection, which is important for robotics perception, as an example for illustration purposes.

\subsection{Input Feature Preprocessing}



\subsubsection{Attentive Region Generator}

For the current input frame $\mathbf{I}$, the attentive regions containing the attentive features are selected by the
Attentive Region Generator (AGG) based on past results.

Three methods can serve this purpose: prediction-based, expansion-based, and hybrid. The prediction-based integrates lightweight prediction to predict the attentive region for the current frame based on the bounding box information from the past frames \cite{yerlikaya2014alternative}. The expansion-based method sets the attentive regions for the current frame by expanding the bounding box from the last frame so that it has high confidence in including the attentive object in the attentive region. The expansion scale can be dynamic based on the characteristics of bounding boxes in past frames. Hybrid methods can combine both prediction and expansion.

\subsubsection{Aggregation Method Optimizer}
Aggregation Method Optimizer determines the methods to aggregate patches or regions optimally. Two options are stitching regions into one input \cite{capel2004image} or processing different regions individually. The second option can also reserve optimality in some cases since we found that twice of the lightweight model inference time may still be faster than the heavyweight model sometimes. 

\subsubsection{Input Feature Aggregation}
This module transfers the input frame $\mathbf{I}$ to $\mathbf{I'}$ based on the attentive region selected and the aggregation method selected. After the Input Feature Preprocessing module, the original input $\mathbf{I}$ is reorganised into a new feature input $\mathbf{I'}$.

\subsection{Efficient Inference with Ensemble of Networks}
In this component, we scale down the full-resolution object detection network to form an ensemble of models with different input sizes. The set of ensembled neural network input sizes is defined as $\mathbb{K}$. The size of the model selected by the Network Selector (NS) is $o = \min(\mathbb{O})$, where $\mathbb{O}$ is the subset of $\mathbb{K}$ larger than the size of $\mathbf{I'}$. Then, $\mathbf{I'}$ is processed by the model $M(o)$ with input size $o$.

\subsection{Result Mapping}
In the last step, the results derived from $\mathbf{I'}$ are mapped back to the global image $\mathbf{I}$. With this step, we obtain the final results in the original input coordinates for other downstream robot tasks.


\section{Experiments}

\subsection{Dataset}
We use the LaSOT dataset \cite{fan2019lasot} as the proof-of-concept validation of our strategy on video object detection of one object to verify the effectiveness of our strategy in dramatically reducing the computation complexity while maintaining accuracy and recall. 

To make the dataset applicable to our testing, we screened all video sequences in LaSOT with the following criteria: (1) The object should also be included in the dataset for training the baseline object detector. In this paper, we use You Only Look Once v7 (YOLOv7) \cite{https://doi.org/10.48550/arxiv.2207.02696} as the baseline object detector, which utilizes the COCO dataset \cite{https://doi.org/10.48550/arxiv.1405.0312} while training. For YOLOv7, we use models W6, E6, and D6 as baseline models; (2) LaSOT is a single object tracking dataset, which only contains at most one bounding box on each frame, but perhaps with more than one target object in the frame. To avoid vagueness and confusion on the object detector, we selected the video sequences with only one object from the target class on each frame. (3) To conduct tests on LaSOT without transfer learning, we selected video sequences where at least 50$\%$ of objects can be detected with the baseline object detector. (4) The minimum edge length of the image is larger than 640. After the dataset screening, 79 video sequences from LaSOT were selected from 18 categories with a total of 243,076 frames. 


\subsection{Implementation Details}
The testing is conducted on a TITAN RTX GPU with 24 Gb memory. We use YOLOv7 object detectors with an input size of 1280 by 1280 as the baseline models for all the testing. For all videos, the ensemble of networks is composed of one default model and three other models downscaled from the default model. The confidence threshold in the model is set as 0.1. The expansion rate for LaSOT in AGG is set to 2. 

\subsection{Metrics}
For the perception performance evaluation, we use the metrics inference time, total time (inference time + time for other overhead computation), Average Recall (AR), average IoU, and average precision AP at IoU of 0.5 and 0.75. AR and AP are calculated according to COCO evaluation protocol, in percentage.

When computing ACP, we use the following $D$: $\alpha \in [-\pi,\pi]$, $|\mathbf{R}_{AB}|\in[0.25 \text{ m},1.5 \text{ m}]$, and $|\mathbf{U}_{AB}|\in[0.02 \text{ m/s},1\text{ m/s}]$. And $T_r=$  0.1 s. When computing CCP, we use the following $C$: $|\mathbf{U}_{AB}|\in[0.02 \text{ m/s},1\text{ m/s}]$, $|\mathbf{R}_{AB}| \in [0.25 \text{ m}, \max(0.25 m, 0.5s\cdot|\mathbf{U}_{AB}|)]$, and $\alpha\in[-\alpha_c,\alpha_{c}]$. 

\subsection{Results on Perception Metrics}
The testing results are summarized in Table \ref{table:lasot result}. Our strategy outperforms by a large margin in both inference time and total time per frame, while only sacrificing a fairly small amount of accuracy on all baseline models. The percentage of reduction in inference time and total time increases with the baseline model complexity. After comparing the inference time and total time for each model, it was found out that the overhead computation time for our strategy and baseline models is almost the same. This demonstrates that the components within our framework shown in Figure \ref{fig:method} are very lightweight and fast to compute. 

\begin{figure}[!t]
\centering
\includegraphics[width=0.9\columnwidth]{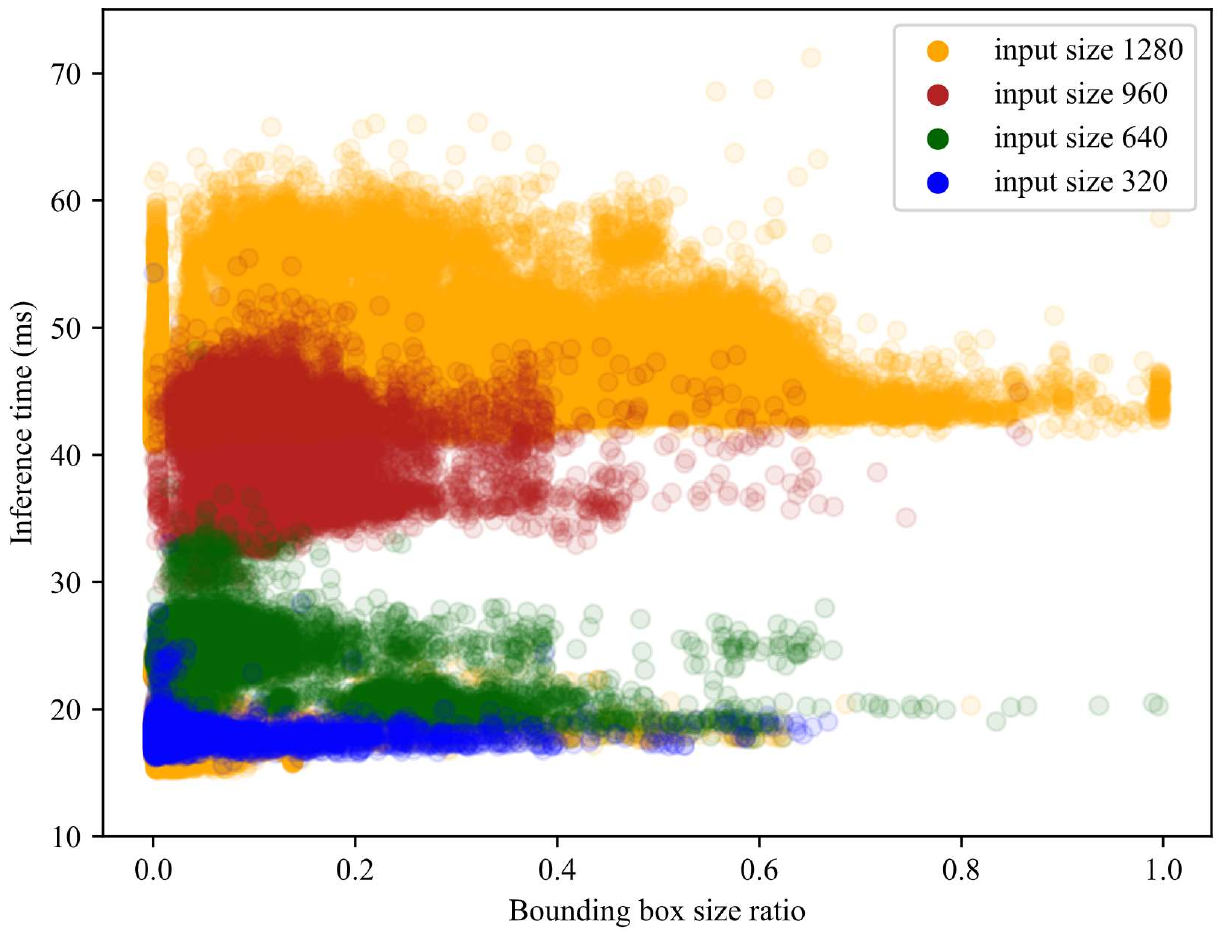}
\caption{Inference time with respect to different bounding box size ratios using models with various input sizes. Each dot represents one frame.  }
\label{fig: size influence}
\end{figure}

\begin{figure*}[!t]
\centering
\includegraphics[width=0.96\linewidth]{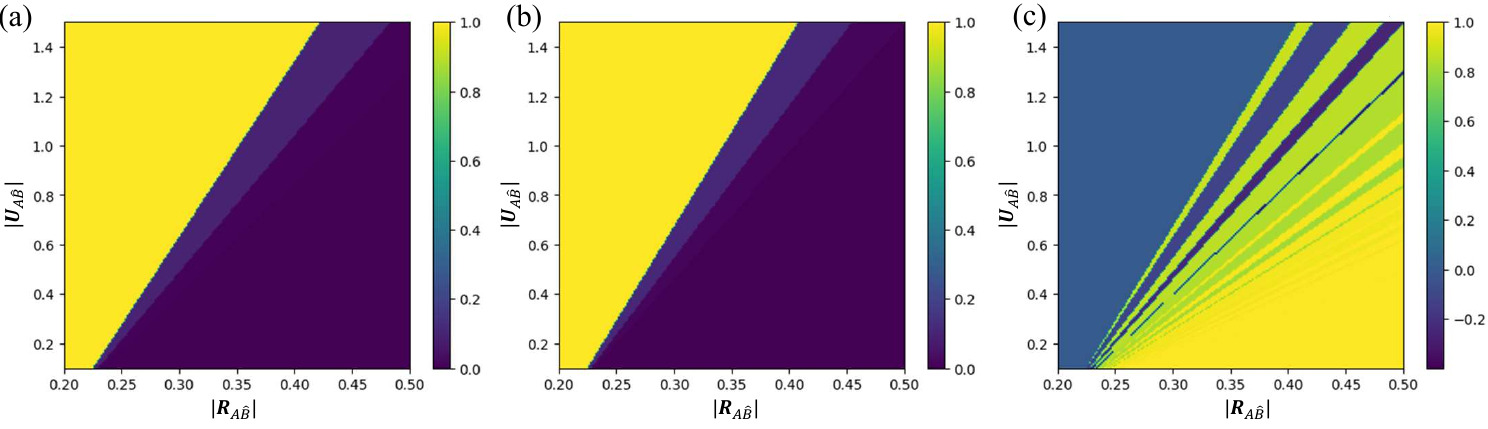}
\caption{Collision probability with respect to distance $|\mathbf{R}_{A\hat{B}}|$ and velocity $|\mathbf{U}_{A\hat{B}}|$ with (a) baseline model; (b) attentive processing strategy. The collision probability decrease percentage is in (c). The values are calculated with E6 model perception metrics from Table \ref{table:lasot result}. In the plot, we set $\alpha =0$. }
\label{fig: result}
\end{figure*}


\begin{table*}[]
\caption{Testing results on LaSOT dataset with three baseline models. $\uparrow$ and $\downarrow$ indicate larger or smaller is better. Testing results verify that our strategy is effective in reducing computation time while maintaining accuracy on all baseline models. }
\centering
\begin{tabular}{cccccccccc}
\hline
\multirow{2}{*}{Metrics} & \multicolumn{3}{c}{YOLOv7-W6}                          & \multicolumn{3}{c}{YOLOv7-E6}                          & \multicolumn{3}{c}{YOLOv7-D6}                          \\ \cline{2-10} 
                         & Baseline        & Ours            & Change \%          & Baseline        & Ours            & Change \%          & Baseline        & Ours            & Change \%          \\ \hline
Inference time $\downarrow$      & 24.377          & \textbf{18.578} & \textbf{-23.792\%} & 34.764          & \textbf{25.036} & \textbf{-27.984\%} & 42.901          & \textbf{29.992} & \textbf{-30.091\%} \\ \hline
total time $\downarrow$         & 30.345          & \textbf{24.465} & \textbf{-19.377\%} & 40.599          & \textbf{30.882} & \textbf{-23.935\%} & 48.791          & \textbf{35.845} & \textbf{-26.534\%} \\ \hline
overhead time $\downarrow$       & 5.967           & 5.887           & -1.340\%           & 5.834           & 5.845           & 0.188\%            & 5.890           & 5.853           & -0.624\%           \\ \hline
$AR$ $\uparrow$                  & \textbf{88.976} & 87.974          & -1.127\%           & \textbf{89.869} & 88.670          & -1.334\%           & \textbf{90.383} & 89.173          & -1.339\%           \\ \hline
IoU $\uparrow$                  & \textbf{0.738}  & 0.729           & -1.218\%           & \textbf{0.750}  & 0.738           & -1.503\%           & \textbf{0.756}  & 0.744           & -1.507\%           \\ \hline
$AP_{0.5}$ $\uparrow$                 & 91.984          & \textbf{92.015} & \textbf{0.034\%}   & \textbf{92.576} & 92.506          & \textbf{-0.076\%}  & \textbf{92.693} & 92.606          & \textbf{-0.094\%}  \\ \hline
$AP_{0.75}$ $\uparrow$                & \textbf{79.391} & 78.952          & -0.553\%           & \textbf{79.939} & 79.487          & -0.566\%           & \textbf{80.446} & 79.938          & -0.632\%           \\ \hline
\end{tabular}

\label{table:lasot result}
\end{table*}

As shown in Figure \ref{fig: size influence}, the average inference time of the neural networks, after scaling down because of the decreased input size, is much shorter. A larger bounding box size ratio (bounding box size over the size of input frame $\mathbf{I}$) with a higher probability corresponds to a larger input size model and, thus, a higher inference time. 

During our testing, the main source of accuracy decrease in this preliminary testing is analyzed. It is found that the smaller model, though fast, is the main source of inaccuracy because of the model scaling in this testing. However, our strategy has a strong potential to overcome this bottleneck and achieve improvement in both computation complexity and accuracy by training from scratch or fine-tuning the lower input size models. The lower input size models only require processing the inputs with objects almost in the center and almost with the same scale because of the cropping operation in the framework. This avoids the location variance and scale variance in object detection. Thus, with these methods, the accuracy has the potential to be improved while maintaining efficiency.



\subsection{Results on the Safety Metrics}
ACP and CCP with various models are summarized in Table \ref{table: safety}. With our proposed attentive processing strategy, the ACP and CCP decrease by a large margin, which verifies the effectiveness in improving safety in human-robot collaborations. Computationally expensive models in the testing are slower to achieve state-of-the-art accuracy and precision on the benchmark. Our strategy works better for those computationally expensive models and reduces CCP and ACP on model D6 by 11.252\% and 13.501\%, respectively. Even though model D6 achieves the highest IoU, AR, and AP as shown in Table \ref{table:lasot result}, its slow processing speed makes model D6 the most dangerous model for HRC as shown in Table \ref{table: safety}. This validates the importance of examining the speed-accuracy-tradeoffs and developing faster real-time algorithms in robotics applications. 




\begin{table}[h]
\centering
\caption{Collision metrics with various models}
\begin{tabular}{c|c|c|c|c}
\hline
   Model                 & Metrics & Baseline   & Attentive Processing & $\mathbf{\downarrow} \%$     \\ \hline
\multirow{2}{*}{W6} & CCP     & 0.461      & $\mathbf{0.431}$                & \textbf{6.508\%}  \\ \cline{2-5} 
                    & ACP     & 5.779$\cdot 10^{-3}$ & $\mathbf{5.359\cdot 10^{-3}}$           & \textbf{7.268\%}  \\ \hline
\multirow{2}{*}{E6} & CCP     & 0.511      & $\mathbf{0.464}$                &   \textbf{9.198\%}     \\ \cline{2-5} 
                    & ACP     & 6.541$\cdot 10^{-3}$ & $\mathbf{5.834\cdot 10^{-3}}$              &  \textbf{10.809\%}      \\ \hline
\multirow{2}{*}{D6} & CCP     & 0.551      & $\mathbf{0.489}$                & \textbf{11.252\%} \\ \cline{2-5} 
                    & ACP     & 7.170$\cdot 10^{-3}$ & $\mathbf{6.202\cdot 10^{-3}}$           & \textbf{13.501\%} \\ \hline
\end{tabular}
\label{table: safety}
\end{table}

The following analysis is based on model E6. The collision probabilities with various $|\mathbf{U_{A\hat{B}}}|$ and $|\mathbf{R_{A\hat{B}}}|$ are shown in Figure \ref{fig: result} (a) and (b). When the distance $|\mathbf{R}_{A\hat{B}}|$ is short with a high relative velocity $|\mathbf{U}_{A\hat{B}}|$, the collision probability is 1 for both baseline and attentive processing. This agrees well with the realistic case since, in these circumstances, the robot doesn't have sufficient time to process the perception or avoid the obstacle. As the velocity decreases or the distance increases, the collision probability decreases in a step function style because the values of $m$ in equation (\ref{eq: m}) are integer and discrete. The collision probability is very small for a large distance and small velocity, which also agrees well with the realistic case. The modeling of collision probability is preliminarily validated.

The collision probability decrease in percentage after adopting our strategy is shown in Figure \ref{fig: result} (c). When the collision probabilities are 1, the change percentage is 0. The main difference between the attentive processing and the baseline is in the computation time $T_p$, recall $AR$, and $IoU$. As the distance increases and/or the velocity decreases, the collision probability with attentive processing decreases from 1 first, which causes a jump in the percentage decrease. The reason for this is the short computation time $T_p$ so that the robot can obtain the information for at least one frame with short $|\mathbf{R}_{A\hat{B}}|$ or large $|\mathbf{U}_{A\hat{B}}|$. However, when the baseline can process the same amount of frames, the collision probability with attentive processing is higher due to the lower accuracy, which is the reason for the negative decrease percentages in some cases. However, these cases become rare when the distance and the velocity further increases and decreases, respectively. Our model of the metric successfully represents this accuracy and speed tradeoff.

\section{Conclusion}
In this paper, we first develop a methodology to understand and model the relationship between robot metrics and three major categories of perception metrics, i.e. detection rate, detection quality, and latency. We further develop and propose two new HRC safety metrics based on perception metrics. To underscore the effectiveness of our model, we develop and propose a novel attentive processing strategy that selects the attentive regions containing the essential components in the input, processes the attentive regions efficiently with an optimal model within an ensemble of networks, and finally maps the results back onto the input frame. 
Our strategy consistently outperforms the baseline models in speed and significantly reduces the total computation time and inference time of an object detector by 26.534\% and 30.091\% while maintaining a similar level of accuracy. Experiments verify that our strategy dramatically reduces CCP and ACP by 11.252\% and 13.501\%, respectively, enhancing HRC safety. With this work, the importance of fast and real-time processing is also demonstrated and verified. Thus, the three concerns we raised in Section I are addressed.



\addtolength{\textheight}{-9.7cm}   




\newpage

\bibliographystyle{IEEEtran}
\bibliography{IEEEabrv,references}

\end{document}